\documentclass{article}

\usepackage{microtype}
\usepackage{graphicx}
\usepackage{booktabs} % for professional tables
\usepackage{adjustbox}
\usepackage{subfig}
\usepackage{listings}
\usepackage{multirow}

\usepackage{hyperref}

\usepackage[accepted]{icml2025}

\usepackage{amsmath}
\usepackage{amssymb}
\usepackage{mathtools}
\usepackage{amsthm}
\usepackage{esvect}
\usepackage[capitalize,noabbrev]{cleveref}

\theoremstyle{plain}
\newtheorem{theorem}{Theorem}[section]

\newtheorem{lemma}[theorem]{Lemma}
\newtheorem{corollary}[theorem]{Corollary}
\theoremstyle{definition}

\theoremstyle{remark}

\usepackage[textsize=tiny]{todonotes}

\icmltitlerunning{Enhancing Attention's Periodic Extension for Length Generalization}

\begin{document}

\twocolumn[
\icmltitle{Fourier Position Embedding:\\Enhancing Attention's Periodic Extension for Length Generalization}




\begin{icmlauthorlist}
\icmlauthor{Ermo Hua}{tsinghua}
\icmlauthor{Che Jiang}{tsinghua}
\icmlauthor{Xingtai Lv}{tsinghua}
\icmlauthor{Kaiyan Zhang}{tsinghua}
\icmlauthor{Youbang Sun}{tsinghua,northeastern}\\
% \icmlauthor{Youbang Sun}{tsinghua}\\
\icmlauthor{Yuchen Fan}{tsinghua,ailab}
\icmlauthor{Xuekai Zhu}{tsinghua,shangjiao}
\icmlauthor{Biqing Qi$^{\dagger}$}{ailab}
\icmlauthor{Ning Ding$^{\dagger}$}{tsinghua}
\icmlauthor{Bowen Zhou$^{\dagger}$}{tsinghua,ailab}\\
\url{https://github.com/TsinghuaC3I/Fourier-Position-Embedding}
\end{icmlauthorlist}

\icmlaffiliation{tsinghua}{Tsinghua University}
\icmlaffiliation{northeastern}{Northeastern University}
% \icmlaffiliation{ailab}{Shanghai AI Laboratory}
\icmlaffiliation{ailab}{Shanghai Artificial Intelligence Laboratory}
\icmlaffiliation{shangjiao}{Shanghai Jiaotong University}

\icmlcorrespondingauthor{Biqing Qi}{qibiqing@pjlab.org.cn}
\icmlcorrespondingauthor{Ning Ding}{dingning@tsinghua.edu.cn}
\icmlcorrespondingauthor{Bowen Zhou}{zhoubowen@tsinghua.edu.cn}

\icmlkeywords{Machine Learning, ICML}

\vskip 0.3in
]



\printAffiliationsAndNotice{}  % leave blank if no need to mention equal contribution

\begin{abstract}
Extending the context length of Language Models (LMs) by improving Rotary Position Embedding (RoPE) has become a trend.
While prior works mainly address RoPE's limitations within attention, this paper uncovers the adverse effects on length generalization from nearly all parts of LMs.
% While prior works mainly address RoPE's limitations within attention, this paper provides an analysis across nearly all parts of LMs, uncovering their adverse effects on length generalization for RoPE-based attention.
Using \textit{Discrete Signal Processing} theory, we show that RoPE enables periodic attention by implicitly achieving \textit{Non-Uniform Discrete Fourier Transform}.
However, this periodicity is undermined by the spectrum damage caused by: 1) linear layers and activation functions; 2) insufficiently trained frequency components brought by time-domain truncation. 
Building on our observations, we propose \textbf{\textit{Fourier Position Embedding (FoPE)}}, which enhances attention's frequency-domain properties to improve both its periodic extension and length generalization. 
FoPE constructs \textit{Fourier Series} and zero-outs the destructive frequency components, increasing model robustness against the spectrum damage.
Experiments across various model scales and benchmarks show that, within varying context windows, FoPE maintains a more stable performance compared to other baselines.
Several analyses and ablations bring further support to our method and theoretical modeling. 
% Code is available at \url{https://github.com/TsinghuaC3I/Fourier-Position-Embedding}.
\end{abstract}

\section{Introduction}

\begin{figure}[h]
    \subfloat[Accuracy on Passkey Retrieval (higher is better)]{
        \includegraphics[width=0.95\linewidth]{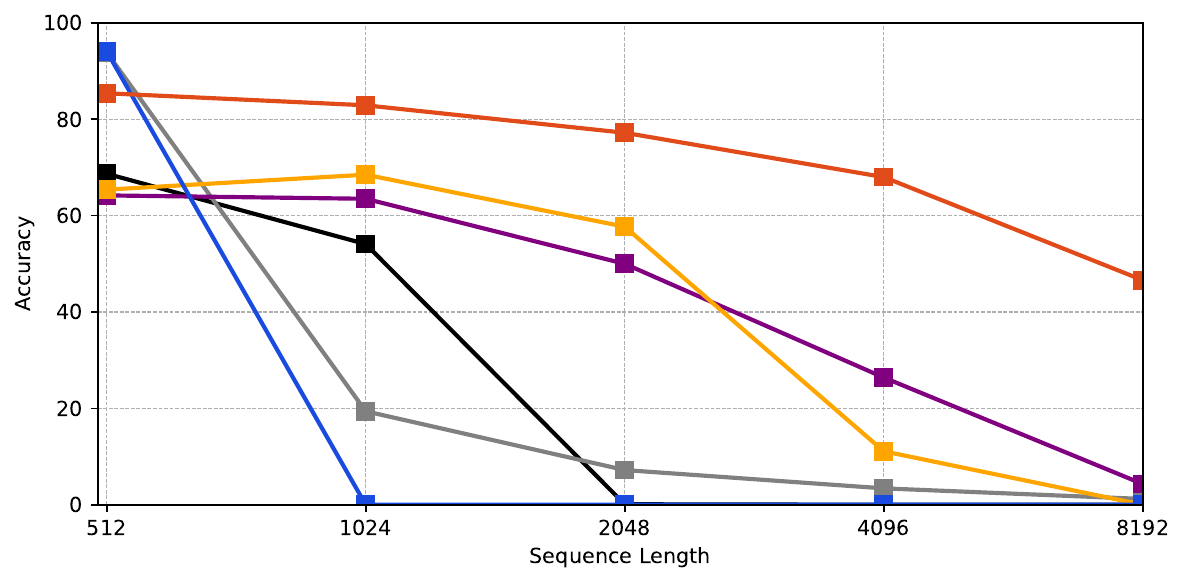}
    }
    \hfill
    \subfloat[Perplexity on C4 (lower is better)]{
        \includegraphics[width=0.95\linewidth]{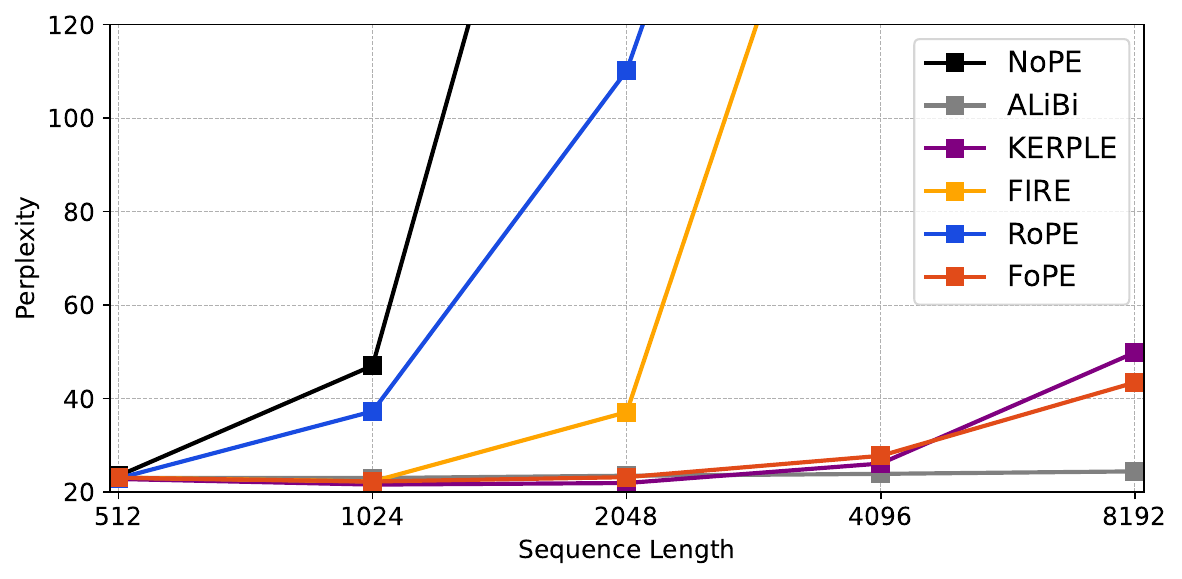}
    }
    \caption{Comparison of 1.2B models with different position embeddings (trained with a maximum sequence length of 512) on a needle-in-haystack task and pre-training perplexity. FoPE yields significantly more stable results than RoPE and other baselines.}
    \label{fig:c4_512}
\end{figure}

% \begin{figure}[h]
%     \subfloat[Accuracy on Passkey Retrieval (higher is better)]{
%         \includegraphics[width=0.9\linewidth]{figures/c4_512_downstream.pdf}
%     }
%     \hfill
%     \subfloat[Perplexity on C4 (lower is better)]{
%         \includegraphics[width=0.9\linewidth]{figures/c4_512_overall.pdf}
%     }
%     \caption{Training with max\_seq\_length = 512, FoPE extends the model's information retrieval ability in a needle-in-haystack task to 16x training length, while performance with RoPE and ALiBi drops dramatically at just 2x training length. When comparing perplexity, FoPE shows significantly more stable performance than RoPE and performs only slightly worse than ALiBi.}
%     \label{fig:c4_512}
% \end{figure}
% However, ALiBi delivers stable perplexity as it considers only the most recent tokens for prediction, lacking true long-context capability.

\begin{figure*}[t]
    \centering
    \includegraphics[width=\textwidth]{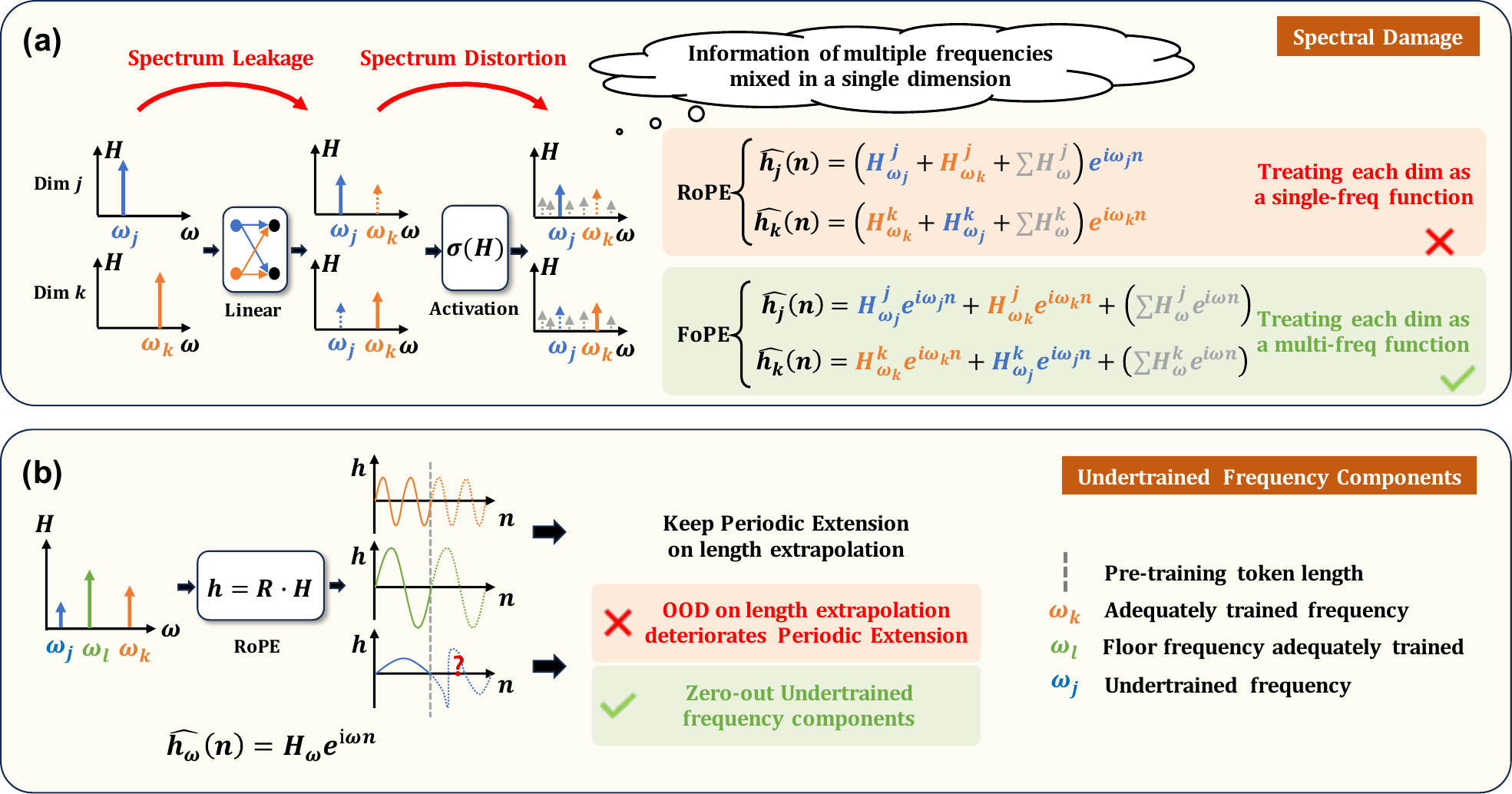}
    \caption{The reasons why RoPE’s periodic extension deteriorates and how FoPE addresses these issues to improve length generalization. (a) As signals pass through linear and nonlinear transformations, this causes spectral leakage and distortion, mixing multiple frequencies into a single dimension. Under RoPE, each dimension is treated as a single-frequency component. By contrast, FoPE models each dimension as a Fourier series of different frequency components, thereby separating information more effectively and mitigating spectral damage. (b) FoPE eliminates inadequately trained frequency components, which are harmful for periodic extension. By preserving only the zero-frequency component, FoPE safeguards periodic extension and delivers more robust length generalization.}
    \label{fig:main}
\end{figure*}

% 长文本能力尤为重要 -> 但受限于计算资源，只能在较小的固定窗长下进行训练 -> 模型会在当前长度过拟合，在面对更长的输入时，模型检索输入中信息进行生成的能力将显著降低 -> 有很多方法解决，其中最基本的方式是引入特殊设计的位置编码
Generation based on the information from long contexts is crucial for Language Models (LMs). However, LMs are typically trained on a fixed context window \cite{vaswani2017attention, touvron2023llama, groeneveld2024olmo} and tends to overfit to the specific context length.

% ALiBi 和 RoPE -> 虽然 RoPE 理论上建模了无限长 Context 的上下文依赖，但仍然会在面对超长文本是丢失信息
Many studies consider the absolute position embedding \cite{vaswani2017attention} as a key contributor to overfitting.
As mitigation, several relative position embedding methods have been proposed \cite{press2021train, chi2022kerple, peng2023yarn, li2024fire, jin2024llm, su2024roformer} to improve LMs' long-distance dependency.
Among these, ALiBi \cite{press2021train} introduced a position-biased attention mask, which linearly declines the attention weights based on distance. ALiBi achieves stable perplexity during pre-training, but fails to retain information from distant tokens, leading to poor performance on long-context downstream tasks.
Another method, RoPE \cite{su2024roformer}, uses the phase of complex numbers to store the position information. Combined with continual pre-training and other interpolation-based methods \cite{peng2023yarn,xiong-etal-2024-effective,chen2024clex,jin2024llm}, RoPE provides better access to long-distance information, making it one of the most widely used position embedding. 
However, RoPE-based LMs still struggle with length generalization without supplementary methods like YARN \cite{peng2023yarn}. 

In this paper, we take a closer look at RoPE in the frequency-domain with tools from \textit{Discrete Signal Processing (DSP)} theory.
Our modeling reveals that RoPE implicitly performs \textit{Non-Uniform Discrete Fourier Transform (NUDFT)} on the hidden states, enabling periodic attention based on the frequency-domain encoding.
However, we find that the periodicity is hindered by the spectral damage caused by: 1) linear layers and activation functions outside attention; 2) inadequately-trained frequency components within attention (See Fig \ref{fig:main}). This explains why RoPE fails to achieve length generalization without assistance from other methods. 

% 为了改进 Attention 的长度外推，我们进一步提出了 Fourier Position Embedding 来改进 Attention 的时移不变性.
Building on our observations above, we propose \textbf{\textit{Fourier Position Embedding (FoPE)}} to further improve the attention's periodic extension for better length generalization. Compared to RoPE, FoPE introduces two main improvements:
1) While RoPE treats each dimension as a single-frequency function, FoPE models each dimension as a \textit{Fourier Series}, consisting of a dominate frequency component and several harmonic components. This approach better mirrors the actual spectrum in LMs and helps attention separate information across different wavelengths, mitigating the negative effects of Spectral Damage.
2) FoPE clips inadequately trained frequency components that is harmful to length generalization. To keep the passing of long wavelength information, we substitute these components with zero, as the zero-frequency component corresponds to the longest wavelength.

We summarize our contribution as follows:

    1. Based on DSP, we provide frequency-domain analysis to reveal the negative influence from nearly all parts from LMs. We find that the length generalization is hindered by the Spectrum Damage arised from: 1) linear layers and activation functions; 2) undertrained frequency components. 

    2. We propose FoPE to improve attention's robustness on the Spectrum Damage. FoPE construct Fourier Series to extract multi-frequency information in each dimension, and clip the frequency of destructive components to zero. Thus, FoPE delivers better periodic extension of attention, thus bringing better length generalizaion.

    3. We conduct experiments across several model scales and datasets. The perplexity in pre-training and the accuracy in needle-in-haystack demonstrate FoPE's superiority over other baselines on length generalization. Evaluation on more complex tasks (i.e. summarization and few-shot question-answering) brings further support to our method and theoretical modeling.

\section{Preliminaries}
\subsection{Non-Uniform Discrete Fourier Transform}
Given a finite sequence of $\{x_n\}:=x_0, x_1, ..., x_{N-1}$ equally-sampled from a continuous function $x$, \textit{\textbf{Discrete Fourier Transform (DFT)}} converts them into equally-spaced components $\{X_m\}:=X_0, X_1, ..., X_{M-1}$ in frequency-domain, the original samples can be recovered by Inverse DFT (IDFT):
\begin{equation}\label{eq:dft}
    \resizebox{0.98\hsize}{!}{%
    $X_m = \sum\limits_{n=0}^{N-1}x_n e^{-i2\pi\frac{n}{N}m}, \ \ x_n = \frac{1}{M}\sum\limits_{m=0}^{M-1}X_m e^{i2\pi\frac{m}{M}n}$
    }
\end{equation}
As $e^{i\omega n}=\cos\omega n+i\sin\omega n$ is periodic in the original domain, DFT implicitly transforms the original function into a linear combination of periodic waves with frequency $\omega_m=2\pi\frac{m}{M}$. 
Thus, DFT is an estimation of the original function, which is lossless only if the original function exactly composes of these specific periodic components. 

To achieve a more accurate approximation, the sampled frequencies ${\omega_m} := \omega_0, \omega_1, \ldots, \omega_{M-1}$ can follow any arbitrary distribution tailored to the data characteristics, subject only to the constraint $\omega_m \in [0, 2\pi)$. This generalized form of DFT is known as the \textit{\textbf{Non-Uniform Discrete Fourier Transform (NUDFT)}}.

\subsection{RoPE implicitly achieves Periodic Attention based on NUDFT}

Given a $M$-dimension query $Q$ and key $K$ for tokens $a$ and $b$, RoPE rotates different dimensions $m$ to different phases:
\begin{equation}
    \widetilde{q_m}(n_a) = Q_me^{i\omega_mn_a}, \widetilde{k_m}(n_b) = K_me^{i\omega_mn_b}
\end{equation}
where $\omega_m=1/{\theta^{^{(2m/M)}}}$ and $\theta$ is the pre-defined parameters in RoPE. Then, the attention weight $h_m(n)$ in each dimension will be calculated as:
\begin{equation}
    \widetilde{h_m}(n) = \widetilde{q_m}(n_a)\widetilde{k_m}^*(n_b) = H_me^{i\omega_mn}
\end{equation}
where $n=n_a-n_b$ and $H_m=Q_mK_m$. Finally, the overall attention weight between different tokens becomes:
\begin{equation}
    h(n) = \sum\limits_{m=0}^{M-1}\widetilde{h_m}(n) = \sum\limits_{m=0}^{M-1}H_me^{i\omega_mn}
\end{equation}
Comparing it with Eq (\ref{eq:dft}), it can be observed that RoPE implicitly achieves a token-level Inverse NUDFT with frequency components $\{\omega_m\}$. 

Based on NUDFT, RoPE models the interactions between different tokens as functions compose of several periodic components, which brings \textbf{\textit{periodic extension}} in each dimension $m$:
\begin{equation}\label{eq:periodic}
    \widetilde{h_m}(n+N_{\omega_m}) =  \widetilde{h_m}(n)
\end{equation}
where $N_{\omega_m}=\frac{2\pi}{\omega_m}$ is this component's period. This property can generalize LMs to longer context.

\section{Spectrum Damage Confine the Length Generalization}
\label{sec:spectrum_damage}
Ideally, RoPE-based Attention achieves periodic extension in any length scenario. However, this extension is confined as a key ideal property that is not guaranteed in LMs.

\begin{figure}[h]
    \centering
    \includegraphics[width=0.95\linewidth]{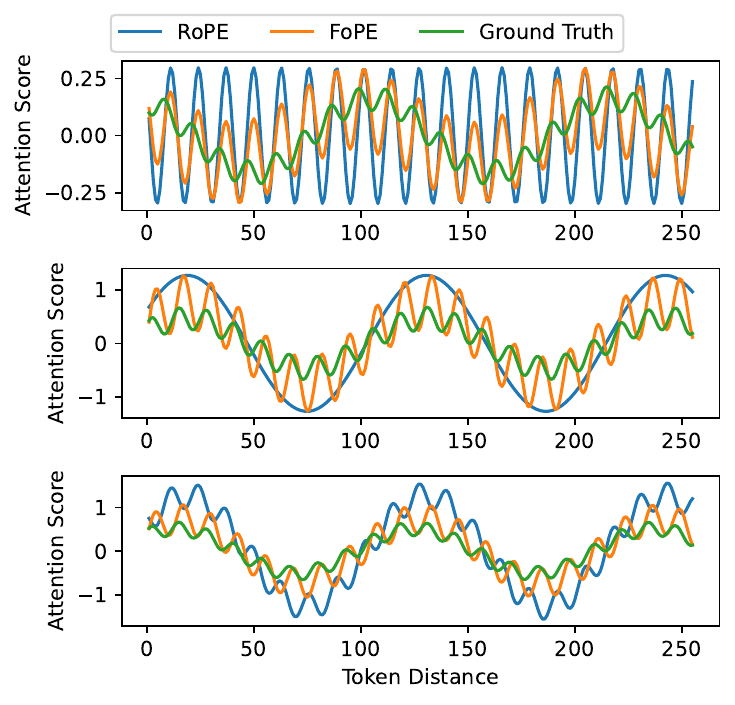}
    \caption{An illustration of the negative impact of Spectrum Damage on length generalization. The figure show the contributions of two frequency components to the final attention score (top \& middle), followed by the sum of their contributions (bottom). We consider a two dimensions hidden states with two frequency components. Because of Spectrum Damage, each dimension actually contains multi-frequency information (referencing Ground Truth). The multi-frequency position embedding introduced by FoPE produces attention scores that are closer to the actual information compared to RoPE.}
    \label{fig:toy}
\end{figure}

\subsection{Negative Influence of Spectrum Damage}
The ideal coefficients and frequencies of NUDFT have one-to-one correspondence. The coefficient of each frequency represents the influence of each token on others propagated at a specific wavelength.

However, the periodic extension is hindered, if the coefficient also contains the information from another frequency component $\omega_o$ with coefficient $H_{\omega_o}=\sigma H_\omega$, called the \textbf{\textit{Spectrum Damage}}. 

If we define the damaged function as $h_{m}' = H_{\omega_m}[(1-\sigma)e^{i\omega_m n}+\sigma e^{i\omega_o n}]$, we find:
\begin{equation}\label{eq:spectrum_damage}
    h_{m}'(n+N_{\omega_m}) \neq h_{m}'(n)
\end{equation}
% \begin{equation}
%     h_{\omega_m,\omega_o}(n+N_{\omega_m}) \neq h_{\omega_m,\omega_o}(n)
% \end{equation}
as $N_{\omega_m}$ is not the period of $h_{\omega_o}$. 
In other words, the information from each component is transmitted through waves with mismatched wavelengths, leading to inaccurate estimation of the influence propagated within each wavelength. (See Fig \ref{fig:toy}) When multiple frequency components are mixed within a dimension, the periodicity of information variation with token distance changes (e.g., Ground Truth in the top and middle sub-figure), no longer matching the periodicity expected by RoPE for that dimension. As a result, the periodic extension and length generalization of attention are adversely affected.

\subsection{Spectrum Damage Outside Attention}
\label{subsec:spectrum_damage}
% Linear Layer 和 Activation Function 带来的频谱损坏
% The establishment of periodic extension has a premise: each coefficient and frequency component need to be one-to-one correspondent. 
% 示意图：线性的频谱泄露 和 非线性的谐波

The LMs' linear layers and activation functions outside attention bring two types of spectrum damage, destroying the one-to-one correspondence between coefficients and frequencies.

% Linear Layer
\textbf{Linear Layer} uses weights $W\in\mathbb{R}^{M \times M}$ to map a $M$ dimension hidden state $X\in\mathbb{R}^M$ to another hidden state $Y\in\mathbb{R}^M$. Thus, each dimension of $Y$ will be a linear combination of different components of $X$:
\begin{equation}
    Y_m = \sum\limits_{k=0}^{M-1}W_{km}X_k
\end{equation}
This results in \textbf{\textit{Spectrum Leakage}}, as different frequency components exhibit interplay. 

% Activation Function

\textbf{Activation Function} has non-linearity in the time domain, generating harmonic frequencies as described by the following Lemma:

% 只有一个频率的情况
% \textit{\textbf{Theorem 1:} Given any single-frequency sinusoid function $\cos\omega n$ and any time-independent non-linear function $g$. The effect of $g$ on $\cos\omega n$ will not change its base frequency and period, but only producing harmonic waves: 
% \begin{equation}
%     g(\cos\omega n) = \sum\limits_{k=0} a_k \cos k \omega n
% \end{equation}
% }

\begin{lemma}
Given a double-frequency sinusoid function $x(n)=\cos\omega_1 n + \cos\omega_2 n$ and any time-independent non-linear function $g$. The effect of $g$ on $x(n)$ yields functions with frequencies as linear combinations of $\omega_1$ and $\omega_2$:
\begin{equation}
    g(x(n)) = \sum\limits_{j \in N}\sum\limits_{k \in N} a_{j,k} \cos(j \omega_1 + k \omega_2)n
\end{equation}
which can be generalized to any multi-frequency function $x(n)=\sum(a_\omega\sin\omega n+b_\omega\cos\omega n)$\footnote{From \cite{oppenheim1982signal}}.
\label{lemma:harmonic}
\end{lemma}

As the hidden states are transformed into multi-frequency functions by Linear Layer, passing them across Activation Functions introduces additional harmonic components, leading to serious \textbf{\textit{Spectrum Distortion}}.

These two types of Spectrum Damage undermine the periodic extension of attention (as shown in Eq.(\ref{eq:periodic})(\ref{eq:spectrum_damage})), hindering the model's length generalization property.

% 消融实验的效果图

\subsection{Spectrum Damage Inside Attention}
% \subsection{Undertrained Frequency Components}
\label{subsec:undertrained_frequency}
% 非充分训练频率的 OOD
Besides the spectrum damage outside attention, the under-trained components of attention within extremely low frequencies ($\omega_m<\frac{2\pi}{N}$) also bring spectrum damage.

Given a single-frequency function $x_m(n)=e^{i\omega_m n}\text{rect}(n)$ truncated by a square wave:
\begin{equation}
    \text{rect}(n)=
    \left\{
        \begin{array}{cc}
            1 & ,n \leq N \\
            0 & ,n > N
        \end{array}
    \right.
\end{equation}
Based on the results of DFT, the spectrum of x(n) is\footnotemark[1]:
\begin{equation}
    X(\omega) = \alpha\delta(\omega_m)+ \frac{\sin[(N-\alpha N_m)(\omega-\omega_m)]}{\omega-\omega_m}
    \label{eq:undertrained}
\end{equation}
where $\alpha=\lfloor\frac{N}{N_m}\rfloor$ and $N_m=\frac{2\pi}{\omega_m}$. 

In the frequency domain, time-domain truncation introduces noisy components via the latter sub-function. When the period of the primary frequency component exceeds the truncation length, its amplitude is significantly weakened. Consequently, noisy components dominate these dimensions, impairing the periodic extension (as defined by Eq.(\ref{eq:periodic})(\ref{eq:spectrum_damage})). 
In contrast, high-frequency components are minimally affected because their coefficients $\alpha$ dominate over the noisy components.

Intuitively, when sampling sinusoidal functions based on token positions, these low-frequency components cannot cover a complete cycle. Therefore, for positions exceeding the pre-training sequence length, these dimensions may sample outside the training domain, leading to difficulties in generalization. Although previous works \cite{peng2023yarn} have identified this issue, we are the first to model it from a Fourier perspective and provide a theoretical explanation.

\begin{figure*}[ht]
    \subfloat[Accuracy on Passkey Retrieval (higher is better)]{
        \includegraphics[width=0.99\textwidth]{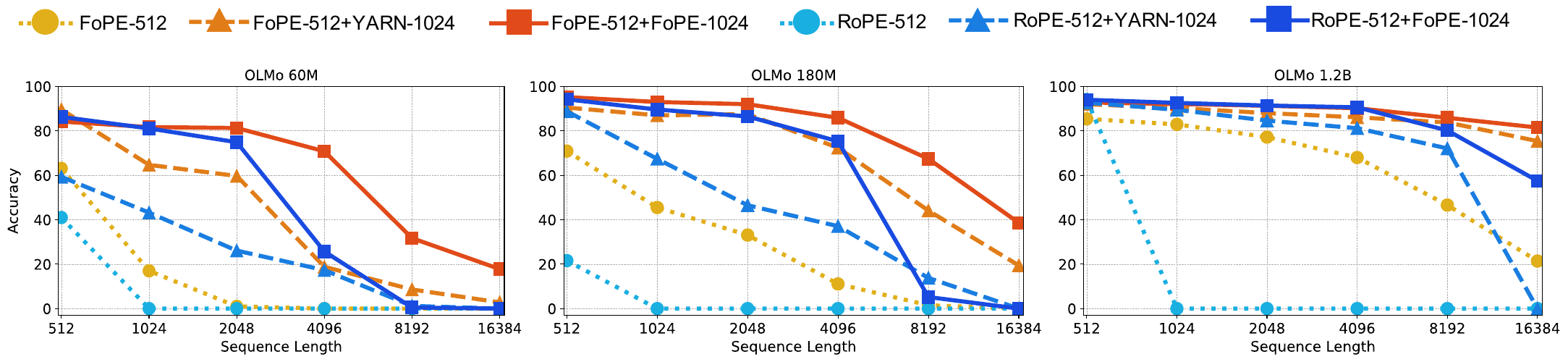}
        % \label{subfig:ablation-sub-methods}
    }
    \hfill
    \subfloat[Perplexity on C4 (lower is better)]{
        \includegraphics[width=0.99\textwidth]{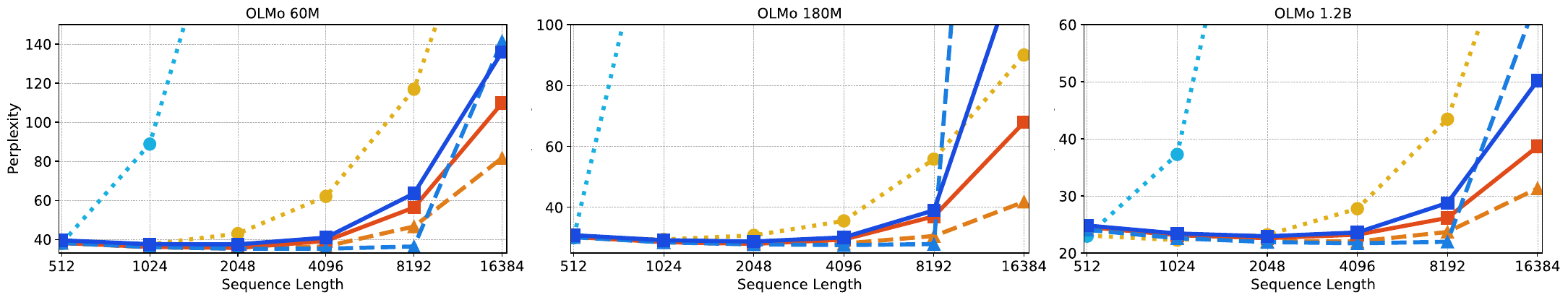}
        % \label{subfig:ablation-sub-methods}
    }
    \caption{Effectiveness of FoPE in length extrapolation. Starting point models trained with 512 context length are extrapolated using YARN and FoPE on a corpus with a maximum sequence length of 1024. The generalization ability of FoPE not only surpasses that of RoPE as a base position embedding, but FoPE also demonstrate capability as an extrapolation method applied to pre-trained models.}
    \label{fig:length-extrapolation}
\end{figure*}

\section{Fourier Position Embedding}
To mitigate the negative affect of the non-ideal frequency-domain properties in LMs, we propose \textit{\textbf{Fourier Position Embedding (FoPE)}} to modify frequency-domain properties of attention:

\textbf{Treating Each Dimension as Multi-Frequency.} Although Linear Layers and Activation Functions bring serious Spectrum Leakage and Spectrum Distortion, they are crucial for enhancing expressive capacity. Therefore, we keep these modules unchanged but focus on modifying how attention processes information within each dimension.

To achieve this, we replace the single frequency in each dimension with Fourier Series:
\begin{equation}
    h_m(n) = H_m(n)(e^{i\omega_m n}+\sum\limits_{\omega}a_{\omega} e^{i\omega n})
\end{equation}
where $a_\omega<1$ because $\omega_m$ is the dominant frequency. This allows attention modules to capture multi-frequency information in each dimension. 

% 还需要写得再清楚一点，比如：omega主要选取已有频率，并说明选择原因
We initialize vector $\{\omega_m\}$ as same as RoPE, and initialize vector $\{\omega\}$ and matrix $\{a_\omega\}$ based on the analysis in Sec \ref{subsec:spectrum_damage}: 
For $\{\omega\}\in\mathbb{R}^D$, we make sure $M \leq D$ so that $\{\omega_m\}\subseteq\{\omega\}$, and the other frequencies can be sampled within $[0, \pi]$ in any distribution.
For $\{a_\omega\}\in\mathbb{R}^{D \times M}$, we initialize it with $N(0,\sigma)$ based on the hypothesis that the Spectrum Damage obeys the similar distribution as the Linear Layers. The coefficients for the real and imaginary part of the frequency are sampled separately in our implementation, which can also use the same coefficient.
The $D$ and $\sigma$ are kept as hyper-parameters to be adjusted.

\textbf{Zero-out Under-trained Frequencies.} As analyzed in Sec \ref{subsec:undertrained_frequency}, the inadequate training of extremely-low frequencies $\omega_m<\frac{2\pi}{N}$ impairs the frequency-domain properties of attention. Thus, we define the floor frequency as $\omega_l=\frac{2\pi}{N}$, and clip the frequencies under the floor frequency to zero. 

We choose zero as the substitute because the zero-frequency component can has the shortest and the longest wavelength at the same time, which presents both long-term and short-term dependency. Also, zero-frequency brings zero average position embedding and will not bring positional bias, which benifits length generalization.

\textbf{Overall function of FoPE} can be formalized as:
\begin{equation}
    h_m(n)=H_m(n)f(\omega_m)
\end{equation}
\begin{equation}
    f(\omega_m) = 
    \left\{
        \begin{array}{cc}
            1 & ,\omega_m < \omega_l \\
            e^{i\omega_m n}+\sum\limits_{\omega}a_{\omega} e^{i\omega n} & ,\omega_m \geq \omega_l
        \end{array}
    \right.
\end{equation}
which treats each dimension either as a Fourier Series or as a zero-frequency component. 

FoPE can be easily achieved with a weight matrix $W^F\in\mathbb{R}^{D \times (M-M_0)}$, where $M_0$ is the number of zero-frequency components in each head (details in Appendix \ref{appendix:implementation})

% \textbf{Implementation of FoPE} can be easily achieved with a weight matrix $W^F\in\mathbb{R}^{D \times (M-M_0)}$, where $M_0$ is the number of zero-frequency components in each head (details in \ref{appendix:implementation}). This matrix maps the coefficients of all frequencies to a Fourier Series for each dimension. 
% Since the zero-frequency sinusoidal function does not affect the original hidden states, the output dimension is less than the dimension of each head. 
% To introduce more diversity and better simulate the randomness of the Spectrum Damage, we assign separate weights for different heads, as well as for the cosine and sine functions. In our implementation, gradients are not required for these matrices, so FoPE adds negligible memory and computation overhead compared to RoPE.

\section{Experiments}
To demonstrate the effectiveness of FoPE as both a position embedding and an extrapolation method, we conduct experiments during pre-training (Sec. \ref{subsec:pre-training}), continual pre-training (Sec. \ref{subsec:continual-pre-training}) and fine-tuning (Sec. \ref{subsec:fine-tuning}). Additionally, we perform ablations to analyze the impact of hyperparameters on FoPE (Sec. \ref{subsec:ablation}), and conduct empirical analysis to showcase the working machenism of FoPE (Sec. \ref{subsec:analysis}). Experimental details and results on more downstream tasks are shown in Appendix \ref{appendix:supple_exp}.

\subsection{Basic Settings}
% TODO: 讲清楚两组setting：① pre-training和continual pre-training；② Fine-tuning；
% TODO: 介绍实验结果
% TODO: Related Work 增加两篇参考文献

\textbf{Pre-training and Continual Pre-training:} In these two settings, we conduct experiments with the OLMo \cite{groeneveld2024olmo} framework and consider different scale models having 60M, 180M, 1.2B parameters. 
We consider two metrics: perplexity for validation set and accuracy on Passkey Retrieval \cite{mohtashami2023landmark}. Passkey Retrieval measures the models' ability in retrieving a short passkey (i.e., a five-digit number) from a large context full of meaningless text. 
We conduct this evaluation based on the implementation from \cite{peng2023yarn}. During evaluation, the passkey is randomly positioned at uniformly distributed locations within the context. For each context length, we test for 1000 trials to ensure the positions sampled are sufficiently dispersed.

\textbf{Fine-tuning:} To assess the effectiveness of our method on more complex downstream tasks, we select SmolLM-1.7B \cite{allal2024SmolLM}, a capable open-source model with the same architecture as Llama \cite{touvron2023llama}, as our base model. We then fine-tune it with the same training recipes as SmolLM-1.7B-Instruct \cite{allal2024SmolLM}. In this setting, we evaluate two types of downstream tasks, summarization and few-shot question-answering. For summarization, we use GovReport \cite{huang2021efficient} and MultiNews \cite{fabbri2019multi}. For few-shot question-answering, we use TREC \cite{li2002learning}, TriviaQA \cite{joshi2017triviaqa}, and SAMSum \cite{gliwa2019samsum}. All these evaluations are conducted under the same setup as in \cite{bai2023longbench}.

\subsection{Length Generalization after Pre-Training}
\label{subsec:pre-training}
We consider two settings to evaluate both the intra-domain and out-of-domain generalization: 

\textbf{Setting 1:} We train models with a 10B-tokens subset of C4 \cite{raffel2020exploring} and evaluate the perplexity in a validation set from C4. 

\textbf{Setting 2:} We train models with $\sim$5B tokens from Gutenberg Books \cite{gutenbergbooks} and evaluate them in the same validation set as Setting 1. In this setting, the language distribution is different between the validation set and the training set, which can further evaluate the generalization ability of different methods.

\begin{figure}[t]
    \centering
    \includegraphics[width=0.9\linewidth]{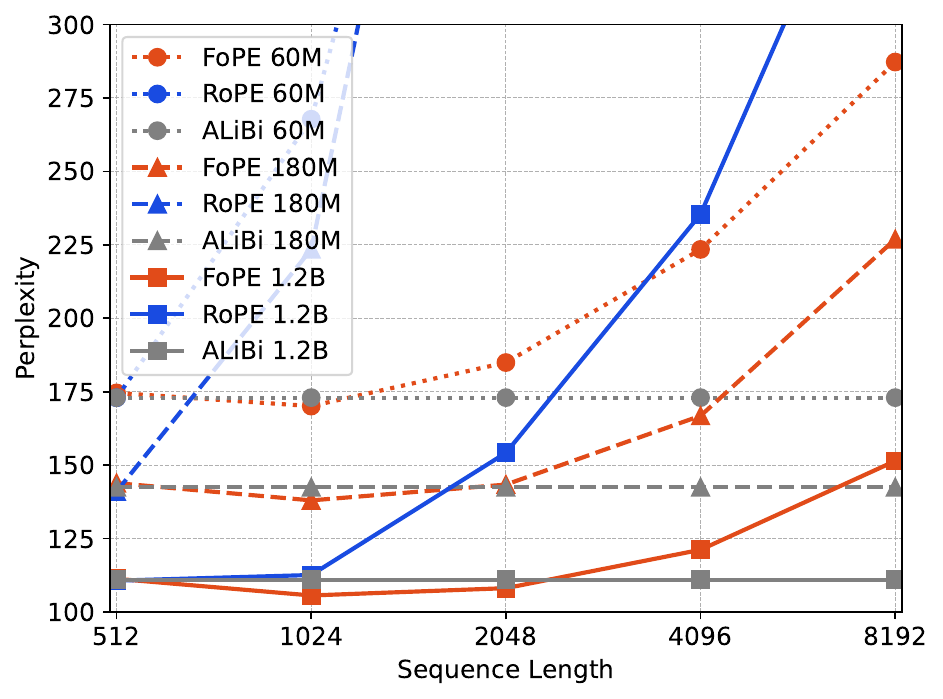}
    \caption{Training with max\_seq\_length=512 on Gutenberg Books and evaluating on a validation set of C4, FoPE also demonstrates its ability to generalize across different data distributions.}
    \label{fig:books_512}
\end{figure}

\textbf{Results of Perplexity.} (See Fig \ref{fig:c4_512}.b \& \ref{fig:books_512}) In both settings, FoPE shows a significant advantage over RoPE. But FoPE is slightly worse than ALiBi, as there is an issue when ALiBi meets this training corpus, which is also mentioned in other papers \cite{peng2023yarn, chen2024clex}. On the one hand, the corpus in C4 and Books mainly have short-distance dependency, thus the information from a short context window is enough for the prediction of almost all tokens. On the other hand, AliBi uses linear declined attention to eliminate long-distance information, and only pays attention to short-distance dependency. Based on these two reasons, ALiBi does not have any decline in perplexity as the context length increases. 

\begin{table}[ht]
    \centering
    \caption{Evaluation on Summarization and Few-shot QA tasks. On most benchmarks, FoPE delivers better length generalization compared to RoPE. The only exception is TriviaQA, where FoPE performs slightly worse in shorter contexts. However, FoPE remains stable up to 8k+ and outperforms RoPE significantly in 8k+.}
    \begin{adjustbox}{width=0.85\linewidth}
        \begin{tabular}{c|c|cl}
            \toprule
            \textbf{Benchmark} & \textbf{Length} & \textbf{RoPE} & \textbf{FoPE}\\
            \midrule
            \multirow{3}{*}{GovReport}
            & 0-4k & 13.02 & 13.27 \color{blue}{$\uparrow 0.25$} \\
            & 4-8k & 11.35 & 12.50 \color{blue}{$\uparrow 1.15$} \\
            & 8k+ & 12.02 & 12.38 \color{blue}{$\uparrow 0.36$} \\
            \midrule
            \multirow{3}{*}{MultiNews}
            & 0-4k & 12.71 & 12.92 \color{blue}{$\uparrow 0.21$}\\
            & 4-8k & 11.11 & 12.98 \color{blue}{$\uparrow 1.87$}\\
            & 8k+ & 10.85 & 12.23 \color{blue}{$\uparrow 1.38$}\\
            \midrule
            \multirow{3}{*}{TREC}
            & 0-4k & 37.00 & 41.00 \color{blue}{$\uparrow 4.00$}\\
            & 4-8k & 42.00 & 56.00 \color{blue}{$\uparrow 14.0$}\\
            & 8k+ & 36.00 & 51.00 \color{blue}{$\uparrow 15.0$}\\
            \midrule
            \multirow{3}{*}{TriviaQA}
            & 0-4k & 36.50 & 33.26 \color{gray}{$\downarrow 3.24$}\\
            & 4-8k & 36.12 & 33.53 \color{gray}{$\downarrow 2.49$}\\
            & 8k+ & 25.02 & 33.87 \color{blue}{$\uparrow 8.85$}\\
            \midrule
            \multirow{3}{*}{SAMSum}
            & 0-4k & 10.27 & 19.77 \color{blue}{$\uparrow 9.50$}\\
            & 4-8k & 6.37 & 15.85 \color{blue}{$\uparrow 9.48$}\\
            & 8k+ & 8.49 & 17.26 \color{blue}{$\uparrow 8.87$}\\
            \bottomrule
        \end{tabular}
    \end{adjustbox}
    \label{tab:smollm_downstream}
\end{table}

\begin{figure*}[t]
    \subfloat[PPL Ratio for different sub-methods]{
        \includegraphics[width=0.31\textwidth]{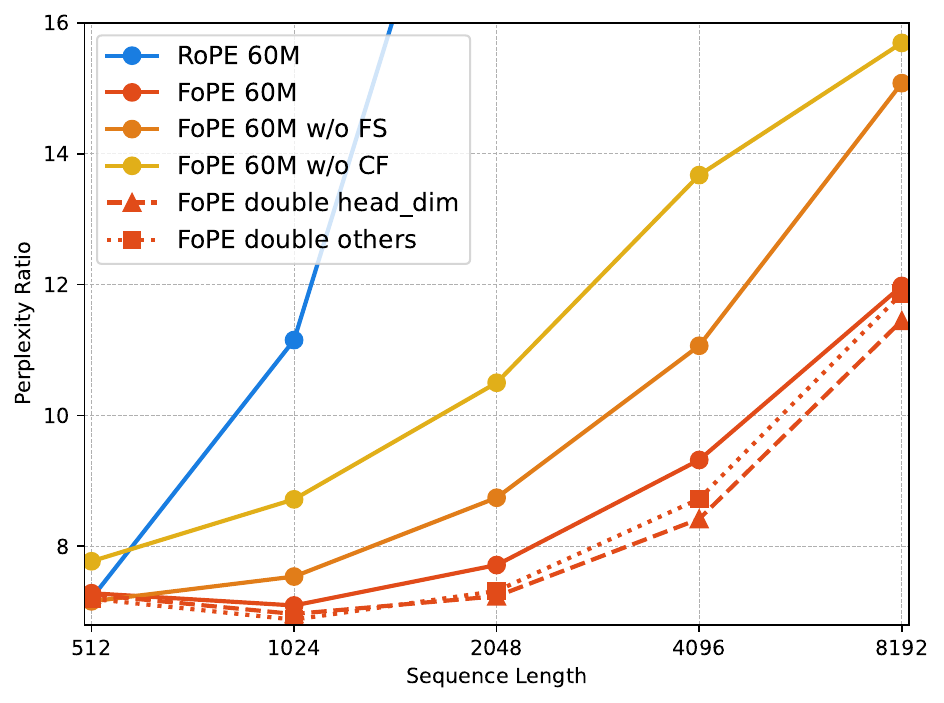}
        \label{subfig:ablation-sub-methods}
    }
    \hfill
    \subfloat[PPL Ratio for different $\sigma$]{
        \includegraphics[width=0.31\textwidth]{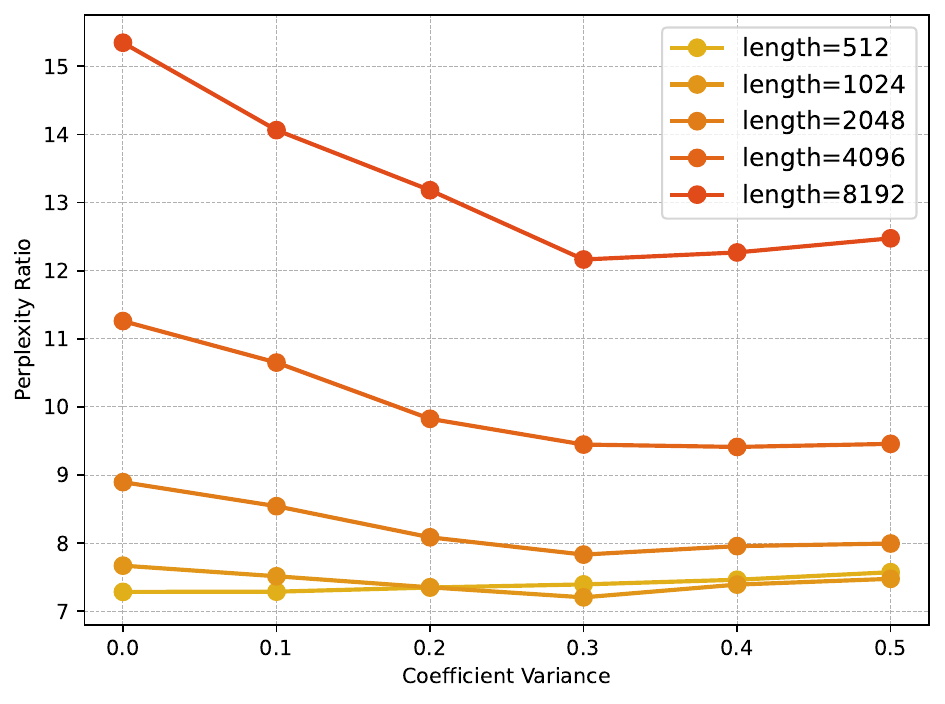}
        \label{subfig:ablation-sigma}
    }
    \hfill
    \subfloat[Passkey Acc for different $D$]{
        \includegraphics[width=0.31\textwidth]{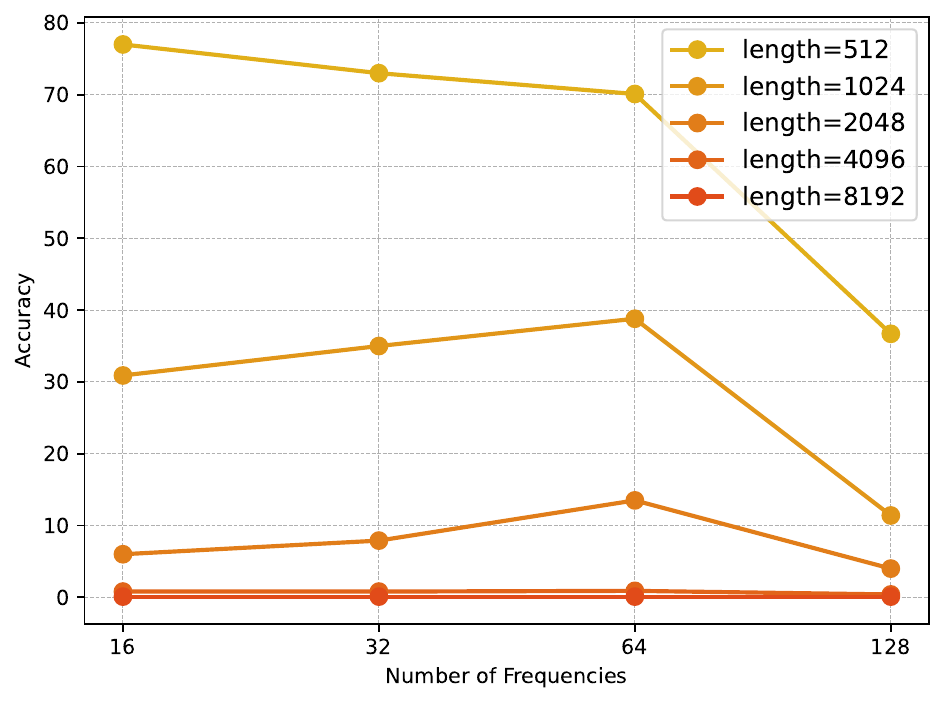}
        \label{subfig:ablation-D}
    }
    \caption{Ablation studies on several hyperparameters. (a) and (b) evaluate $\text{PPL Ratio}=\text{PPL}_{\text{c4}}/\text{PPL}_{\text{books}}$, while (c) evaluates accuracy on Passkey Retrieval as we found $D$ slightly influence the perplexities of training. See the detailed analysis in Sec \ref{subsec:ablation}.}
\end{figure*}

% \subsection{Passkey Retrieval Task}
% \label{subsec:passkey}
\textbf{Results of Passkey.} (See Fig \ref{fig:c4_512}.a) FoPE demonstrates a significant advantage over all baselines in this task. RoPE’s accuracy drops sharply to zero at twice the training length and remains at zero for longer sequences. ALiBi shows a linear decline in accuracy, further illustrating that its linearly declining attention is unable to capture information from long distances. In contrast, FoPE maintains stable retrieval accuracy at any position, demonstrating a strong ability to extract subtle information from long sequences.

\subsection{Length Generalization after Continual Pre-Training}
\label{subsec:continual-pre-training}
Beyond the use of positional embeddings during the original pre-training phase, several extrapolation methods \cite{peng2023yarn, chen2024clex} have been proven critical for enhancing length generalization. Thus, we investigate two key aspects of FoPE: 1) whether existing extrapolation methods are also effective for FoPE; 2) whether FoPE can enable extrapolation on RoPE-based models, thereby allowing seamless integration with existing open-source models. In this sub-experiment, we select a representative extrapolation method, YARN \cite{peng2023yarn}, as our baseline. We fine-tune the last checkpoint from pre-training for $\sim$ 1B tokens in this setting.

\textbf{Results.} (See Fig \ref{fig:length-extrapolation}) In comparison with RoPE+YARN, FoPE+YARN achieves significantly better length generalization performance, as demonstrated by lower perplexity on the C4 dataset and higher accuracy in the Passkey Retrieval task. Moreover, FoPE outperforms YARN in length extrapolation for both RoPE-based and FoPE-based models. These findings underscore the effectiveness and practical utility of FoPE, which holds the potential to enhance all RoPE-based open-source models.

\subsection{Length Generalization after Fine-Tuning}
\label{subsec:fine-tuning}
We further evaluate the effectiveness of FoPE within the widely used LLaMA architecture, focusing on more challenging tasks such as summarization and few-shot question answering. For this purpose, we select SmolLM-1.7B, which is pre-trained with a context length of 2048. We then fine-tune SmolLM to a context length of 4096 using both RoPE and FoPE, and compare their performance on samples of varying sequence lengths from selected datasets, categorized into 0–4k, 4–8k, and 8k+.

\textbf{Results.} (See Table \ref{tab:smollm_downstream}) On nearly all benchmarks, FoPE not only improves inner-domain performance for samples with 0–4k length, but also demonstrates more stable generalization capabilities for longer context lengths. The only exception occurs in TriviaQA, where FoPE performs slightly worse in shorter contexts. However, FoPE’s performance remains stable up to 8k+ and significantly outperforms RoPE.

\subsection{Ablation Studies}
\label{subsec:ablation}
Considering the consistent performance of FoPE across different parameter scales and datasets, we conduct ablation studies only on the 60M models trained on 5B tokens from Gutenberg Books. Except for the primary ablation variable, all other hyper-parameters are kept identical to the main experimental settings.

\textbf{Both sub-methods of FoPE are useful} (See Fig \ref{subfig:ablation-sub-methods}). FoPE is constitutive of two parts, called \textit{Fourier Series (FS)} and \textit{Clip Floor to Zero (CF)}. Although these two sub-methods are both useful for length generalization, combining them together brings a more significant improvement. On one hand, FS contributes more to length generalization, which demonstrates that the Spectrum Damage have a significant influence on length generalization. On the other hand, CF contributes more to fitting the current dataset and sequence length, which implies the zero-frequency component is the most informative and indispensable component.

% \textbf{Does the threshold for clipping frequency need to be the Floor Frequency $\frac{2\pi}{N}$?}

\textbf{Increasing the dimension of attention heads is more beneficial than increasing the number of attention heads or layers} (See Fig \ref{subfig:ablation-sub-methods}). More dimensions introduce more frequency components, making attention more robust to Spectral Damage. In contrast, adding more attention heads and layers aggravates Spectrum Damage, which diminishes the benefits of expanding the parameter scale.

\textbf{Variance $\sigma$ of $\{a_\omega\}$} (See Fig \ref{subfig:ablation-sigma}). We keep $D=16$ to only evaluate $\sigma$'s influence. By grid searching $\sigma$ from 0 to 0.5, we find that setting $\sigma=0.3$ for 60M model obtain the best perplexity, especially for longer context. The best $\sigma$ implies the estimated strength of Spectrum Damage of the 60M model, and the estimation may become larger as the models' parameter scale increases.

\textbf{Number $D$ of $\{\omega\}$.} We keep $\sigma=0.3$ to only evaluate $D$'s influence. By grid searching $\sigma$ from 16 to 128, we find that $D$ does not significantly influence the perplexity, but it is important for Passkey Retrieval. Setting $D=64$ can obtain the best accuracy for Passkey Retrieval. The best $D$ is the estimated number of strong enough noisy components of each model, and this number may become larger as the parameter scale increases. The harmonic frequencies tend to be weaker than the base frequencies, and this phenomenon is more significant to the higher-order harmonics. Thus, there are limited noisy frequency components that have enough intensity to disturb the passing of the base wave, paying attention to not important components hinders the effectiveness of the model.

Although the choice of hyper-parameters affects the effectiveness of FoPE, the ablation study shows that FoPE consistently outperforms RoPE across all hyperparameter settings, demonstrating its robustness to hyperparameter selection.

\begin{figure}[ht]
    \centering
    \subfloat[Q vectors' mean activation across heads]{
        \includegraphics[width=\linewidth]{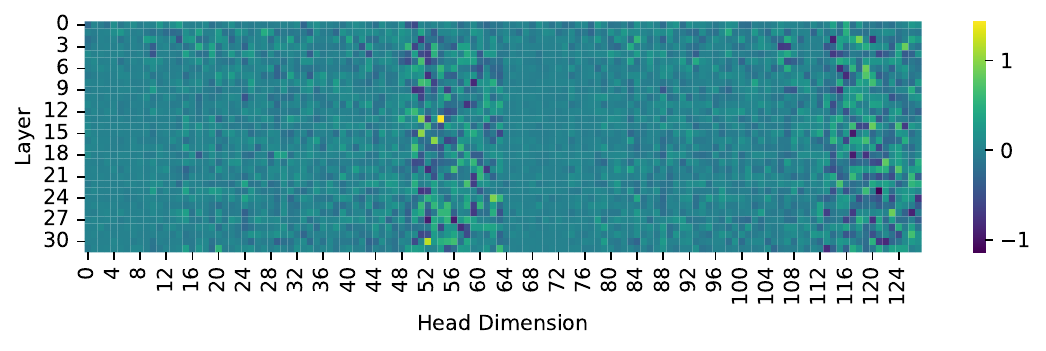}
    }
    \hfill
    \subfloat[K vectors' mean activation across heads]{
        \includegraphics[width=\linewidth]{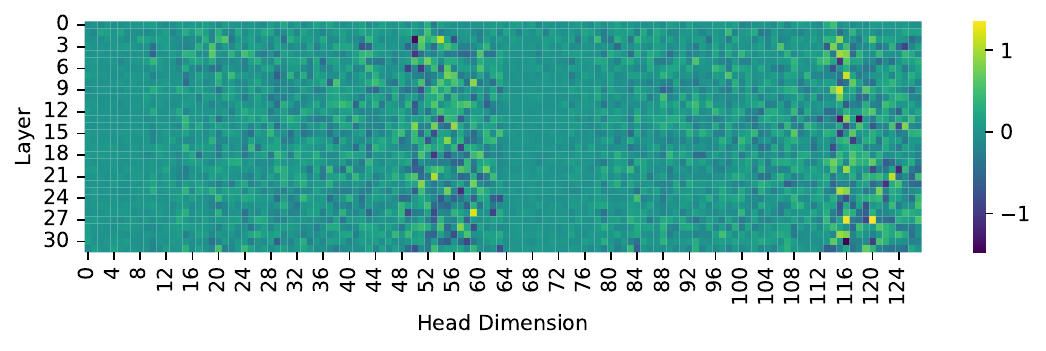}
    }
    \caption{The activation values across different dimensions of $q$, $k$ vectors in various layers, averaged over the attention heads in Llama2-7B. We found that under-trained dimensions holds higher absolute values across all layers of the model.}
    \label{fig:qk}
\end{figure}

\begin{table}[ht]
    \centering
    \caption{The loss of 20M toy models trained on a sequence length of 512. We consider three types of position embeddings, among which only RoPE has frequencies that cannot complete full cycles. For RoPE-A, all frequencies are adjusted from RoPE to the nearest values that exactly complete full cycles.}
    \begin{adjustbox}{width=0.48\textwidth}
    \begin{tabular}{l|c|c|c|c|c}
        \toprule
        \textbf{Sequence Length} & \textbf{512} & \textbf{1024} & \textbf{2048} & \textbf{4096} & \textbf{8192} \\
        \midrule
        RoPE & 5.50 & 6.01 & 6.58 & 6.99 & 7.16\\
        RoPE + QK\_Norm & \underline{5.46} & \underline{5.56} & \underline{5.89} & \underline{6.32} & \underline{6.66} \\
        \midrule
        RoPE-A & 5.72 & \underline{5.86} & \underline{6.18} & \underline{6.46} & \underline{6.67}\\
        RoPE-A + QK\_Norm & \underline{5.69} & 5.89 & 6.27 & 6.59 & 6.81 \\
        \midrule
        NoPE & 5.65 & \underline{6.03} & \underline{6.60} & \underline{6.81} & \underline{6.99} \\
        NoPE + QK\_Norm & \underline{5.59} & 6.18 & 6.87 & 7.10 & 7.43 \\
        \bottomrule
        \end{tabular}
    \end{adjustbox}
    \label{tab:qk_norm}
\end{table}

\subsection{Empirically Validation of FoPE's Mechanism}
% \subsection{The Necessity to Zero-Out Undertrained Components}
\label{subsec:analysis}

To further validate the theoretical insights of FoPE analyzed before, we conduct empirical experiments on its two core modification of RoPE.

\textbf{Modeling each dimension as Fourier Series.} 
We use a toy model consisting of a single-layer MLP with an activation function, to validate the necessity of multi-frequency representation in a single dimension by simulating attention scores' calculation. For simplicity, we model the interaction of a token with only two dimensions (frequency components) hidden states, setting both Query and Key matrices as identity matrices during attention score computation. We track the transformations of each dimension through the MLP and activation function, matching frequency components with corresponding sinusoidal functions to obtain frequency-matched attention scores, termed Ground Truth (See Fig \ref{fig:toy}).  Comparing Ground Truth with RoPE and FoPE attention scores, we find that FoPE more accurately captures multi-frequency periodicity, yielding attention scores that better align with information transfer.

\textbf{The negative effect of under-trained dimensions.} 
By visualizing the numerical expectations of the $q$ and $k$ vectors in each dimension (details in Appendix \ref{appendix:visualize-of-qk}), we observe that the absolute values of the dimensions corresponding to under-trained frequencies holds higher absolute values across all layers of the model. This positional bias may adversely affect robustness to out-of-domain rotation matrix values during length generalization. To further verify this hypothesis, we train several toy models (See  Table \ref{tab:qk_norm}). For models with QK\_Norm, we normalize the $q$,$k$ vectors (enforcing a mean of 0 and variance of 1) before applying rotation matrix to eliminate positional bias. Such normalization delivers positive impacts only on under-trained frequency components, but not on frequency components complete full cycles. These experimental results validate our hypothesis.

\section{Related Work}
\textbf{Frequency-Domain Embedding.}
% 傅里叶变换 -> 傅里叶变换在通信、语音信号处理等领域的应用（因为空间中的信号自带周期性和时移不变性）-> 傅里叶变换在机器学习中相关研究
Discrete Fourier Transform (DFT) \cite{oppenheim1982signal} has been widely used in various areas having periodic signals \cite{edfors2000analysis, sanchez2010cluster}.
In machine learning, \cite{uteuliyeva2020fourier, lin2024cyclenet, tancik2020fourier, tamkin2020language, lee2022fnet, gillman2024fourier} employed Fourier features into neural networks to enhance performance on NLP or CV tasks.
S4 \cite{gu2021efficiently} also leveraged FFT and IFFT to shift its core computation into the frequency-domain, delivering more efficient computation.
\cite{wang2019encoding, su2024roformer} improved the attention mechanism by defining position embedding with complex number, while "phase" used in these methods is a typical concept in frequency-domain.

\textbf{Length Generalization.}
% 训练资源不足，需要长度外推 -> 位置编码 -> 调整位置编码，直接外推的方法
Due to resource constraints, LMs are trained on limited-length corpus chunks and struggle with longer contexts \cite{voita2023neurons, dong2024exploring, hong2024token, qi2024smr}.
While absolute position embeddings \cite{vaswani2017attention} restrict the general use of positional information, methods as \cite{shaw2018self, yang2019xlnet} directly adjust the attention mechanism, 
another intuitive method is to redesign the position embedding \cite{press2021train, chi2022kerple, li2024fire, kazemnejad2024impact, su2024roformer, wang2024length, choromanski2024learning, barbero2024round, chen2024hope}. 
Among these, RoPE \cite{su2024roformer} encodes positional information using the phase of complex numbers, leveraging their periodicity to enhance access to long-distance dependencies.
Several training-free or fine-tuning-based methods can also improve the LM's length generalization by refining RoPE \cite{peng2023yarn,chen2024clex,jin2024llm,lin2024mixture}. However, these works mainly address the drawbacks of RoPE in attention mechanism, neglecting the influence of other components in LMs.

\section{Conclusion}

In this paper, we analyze RoPE-based attention by modeling it in the frequency domain using \textit{Discrete Signal Processing (DSP)} theory. 
Our analysis reveals that RoPE achieves periodic attention by implicitly performing \textit{Non-Uniform Discrete Fourier Transform (NUDFT)}. However, this periodicity is corrupted by the non-ideal spectrum properties introduced by several parts in LMs.
% Our analysis reveals that RoPE achieves periodic attention by implicitly performing \textit{Non-Uniform Discrete Fourier Transform (NUDFT)}, corrupted by the non-ideal spectrum properties brought by other parts in LMs.
We propose \textbf{\textit{Fourier Position Embedding (FoPE)}} to enhances attention's periodic extension and length generalization. FoPE models each dimension as Fourier Series and zero-out inadequately-trained frequency components. 
% Experiments demonstrate that FoPE significantly improves length generalization compared to baselines across diverse tasks and datasets. 
Experiments are conducted on models with various architectures and scales. Based on the evaluation on diverse well-known datasets including summarization and question-answering tasks, FoPE demonstrates significantly improvement length generalization compared to baselines.
% Our ablation studies provide further support for our method and theoretical modeling.
Our ablation studies brings analyze on the influence of each sub-method and hyperparameter on FoPE. Several empirical toy experiments also provide further support for our method and theoretical modeling.

\section*{Impact Statement}

Our DSP-based modeling in the frequency domain provides a novel perspective for LMs to enhance length generalization and explore broader applications.
These include aligning frequency-domain representations for better model collaboration, optimizing kv-cache compression via spectral analysis, and improving semantic communication with learnable frequency-domain embeddings.

\section*{Acknowledgement}

This work is supported by the National Science and Technology Major Project (2023ZD0121403), the Beijing Natural Science Foundation (IS23059), the Young Elite Scientists Sponsorship Program by CAST (2023QNRC001), and
the National Natural Science Foundation of China (No. 62406165). We further extend our gratitude to Yihao Liu, Yiming Shi, Yizhou Jiang, Hang Deng and Chuxuan Shan, for their insightful discussion with us. 


\bibliography{example_paper}
\bibliographystyle{icml2025}

%%%%%%%%%%%%%%%%%%%%%%%%%%%%%%%%%%%%%%%%%%%%%%%%%%%%%%%%%%%%%%%%%%%%%%%%%%%%%%%
%%%%%%%%%%%%%%%%%%%%%%%%%%%%%%%%%%%%%%%%%%%%%%%%%%%%%%%%%%%%%%%%%%%%%%%%%%%%%%%
% APPENDIX
%%%%%%%%%%%%%%%%%%%%%%%%%%%%%%%%%%%%%%%%%%%%%%%%%%%%%%%%%%%%%%%%%%%%%%%%%%%%%%%
%%%%%%%%%%%%%%%%%%%%%%%%%%%%%%%%%%%%%%%%%%%%%%%%%%%%%%%%%%%%%%%%%%%%%%%%%%%%%%%
\newpage
\appendix
\onecolumn

\section{More Experimental Results}
\label{sec:appendix}

\subsection{Visualization of $q$,$k$ vectors before applying RoPE}
\label{appendix:visualize-of-qk}
We conduct visualization experiments using the Llama2-7B model, which features an attention module with 32 heads, each comprising 128 dimensions, and a pretraining sequence length of 4096 tokens. The number of cycles sampled by each sinusoidal function during pretraining is calculated as $r_i=\frac{\theta_iL_{train}}{2\pi}$, where $i$ denotes the dimension index. Based on this, we determined that the dimensions corresponding to incomplete cycles fall within the ranges $[45,64]\cup [109,128]$.

We randomly sample 1000 tokens and compute the average activation values across heads for each dimension. The average activation values for every dimension of each layer are then plotted in Fig \ref{fig:qk}. The absolute activation values are significantly higher for dimensions corresponding to undertrained frequencies compared to others. This indicates that the RoPE rotation matrix pattern during pretraining has a notable impact on the distribution of $q$,$k$ vector activations.

\subsection{Influence of undertrained components in time-domain}

We conduct visualization and ablations to further investigate the influence of undertrained components in RoPE. 

\begin{figure}[h]
    \centering
    \includegraphics[width=0.5\linewidth]{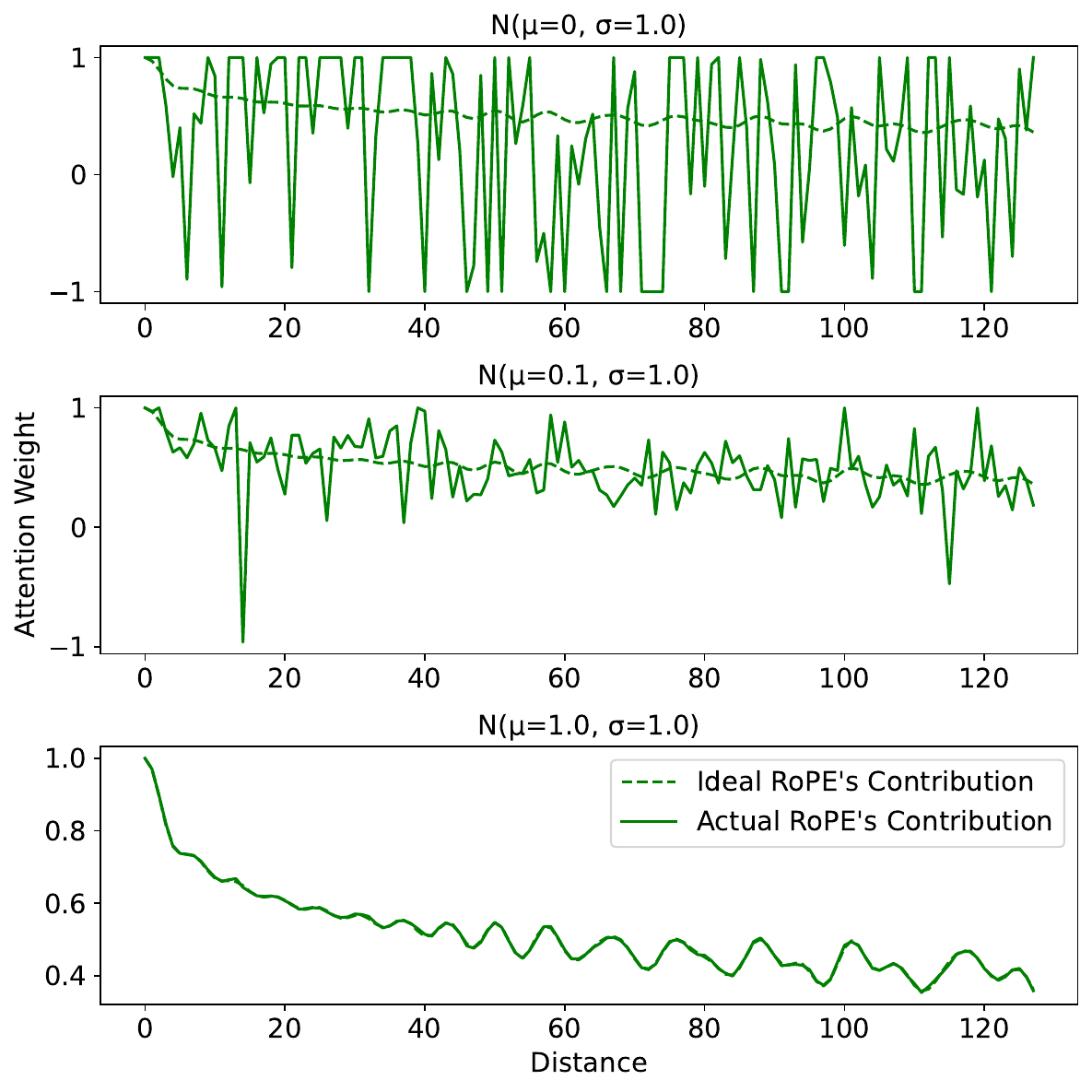}
    \caption{The statistical average contribution of RoPE to attention scores. We sample 1k Q and K vectors from Gaussian distributions with mean of 0, 0.1, and 1.0. The long-distance decay effect weakens as the mean decreases, disappearing entirely when the mean is 0.}
    \label{fig:decay}
\end{figure}

In Fig \ref{fig:decay}, we visualize the time-domain pattern of RoPE. To get the "Ideal RoPE's Contribution", we suppose the Query and Key vectors are equal to 1 constantly (as in the original paper of RoPE \citep{su2024roformer}. To get the "Actual RoPE's Contribution", we suppose the Query and Key vectors obey the Gaussian distribution and sample 1000 times to get the mathematical expectation of the "Actual RoPE's Contribution". It can be seen that RoPE brings decay in attention score in its naive setting, which is brought by the undertrained components in RoPE. We hypothesis that this decay brings positional bias that may adversely affect robustness to out-of-domain rotation matrix values during length generalization. 

As the positional bias can be eased if the mean of Query and Key vectors is set to zero, we conduct further ablation studies to normalize the Query and Key vectors before attention. Based on the results in \ref{tab:qk_norm}, only if the position embedding contains components that cannot complete full cycle, the normalization brings better length generalization. Thus, these components are partially proved to deliver negative affect by positional bias. This experiment also demonstrate that the positional decay does not have beneficial influence on length generalization.

% \subsection{Visualization of FoPE's attention pattern}
% 同时画出 Time Domain 和 Frequency Domain，分上子图和下子图
\subsection{Setups of main experiments}
\label{appendix:main_exp}
Our main experiments are conducted with 4 cards NVIDIA A6000 (maximum GPU memory=48GB). 

% \begin{figure}[ht]
%     \subfloat[Accuracy on Passkey Retrieval (higher is better)]{
%         \includegraphics[width=\linewidth]{figures/c4_512_downstream.pdf}
%     }
%     \hfill
%     \subfloat[Perplexity on C4 (lower is better)]{
%         \includegraphics[width=\linewidth]{figures/c4_512_overall.pdf}
%     }
%     \caption{Trained with max\_seq\_length = 512 on C4, FoPE outperforms RoPE and ALiBi across scales and shows better scalability, with larger models exhibiting slower performance degradation at longer context lengths.}
%     \label{fig:c4_512}
% \end{figure}

\textbf{Pre-Training Settings.} The pre-training of 60M/180M/1.2B OLMo in 10B tokens lasts for 10/20/100 hours, respectively. The time-consuming has a linear relation with the number of tokens in other settings. For all model scales and experimental settings, we select 6e-4 as the learning rate and warm-up for 10000 steps with cosine scheduler. While the mini-batchsize on each device is different for each model, we accumulate gradients until the global batchsize reaches 1024 in all experiments.

\textbf{Fine-Tuning Settings.} Our fine-tuning procedure totally aligns with the official SmolLM-1.7B-Instruct recipe \cite{allal2024SmolLM}. We fine-tune SmolLM-1.7B with approximately 350k samples for one epoch, using the AdamW optimizer with a learning rate of 3e-4 and a cosine scheduler with a warmup ratio of 0.1. The entire training process takes 4 hours on our machines.

\begin{table}[h]
    \centering
    \caption{The hyper-parameter of different model scales. The upper part is the parameter shared by all experiments, while the lower part is the parameter specific to FoPE. We use different mini-batch sizes depending on the max\_seq\_length: the first corresponds to max\_seq\_length=512, and the second corresponds to max\_seq\_length=1024.}
    \begin{adjustbox}{width=0.3\textwidth}
        \begin{tabular}{l|ccc}
            \toprule
            \textbf{Model Scale} & 60M & 180M & 1.2B \\
            \midrule
            \textbf{Num\_Heads} & 8 & 8 & 16 \\
            \textbf{Num\_Layers} & 8 & 8 & 16 \\
            \textbf{MLP\_Ratio} & 8 & 8 & 8 \\
            \textbf{Head\_Dim} & 64 & 128 & 128 \\
            \textbf{Mini\_Batchsize} & 64/32 & 32/16 & 8/4 \\
            \midrule
            \textbf{Var\_Freq $\sigma$} & 0.3 & 0.4 & 0.6 \\
            \textbf{Num\_Freq $D$} & 64 & 128 & 128 \\
            \bottomrule
        \end{tabular}
    \end{adjustbox}
    \label{tab:hyper}
\end{table}

\textbf{Evaluation Settings.} For pre-training, evaluations are conducted on the checkpoint from the last step. For fine-tuning, we save checkpoints every 100 steps and report the best result for each method. This is partly due to YARN \citep{peng2023yarn} being prone to overfitting with excessive fine-tuning steps, a limitation not observed in FoPE.

\subsection{Experimental results on more benchmarks}
\label{appendix:supple_exp}
In addition to evaluating the long-context capabilities of FoPE, we also assessed its performance on a variety of other tasks (see Tables \ref{tab:downstream_acc}, \ref{tab:downsream_loss}, and \ref{tab:mmlu}). The results show that FoPE performs comparably to, or even better than, other baselines, demonstrating that FoPE can enhance long-context capabilities without compromising performance on other tasks.

\begin{table*}[h]
    \centering
    \caption{Accuracy of 1.2B models with different positional embeddings on several downstream tasks.}
    \begin{adjustbox}{width=\textwidth}
    \begin{tabular}{l|ccccccccc|c}
        \toprule
        \textbf{Method} & \textbf{basic arithmetic} & \textbf{social iqa} & \textbf{winogrande} & \textbf{openbook qa} & \textbf{sciq} & \textbf{hellaswag} & \textbf{piqa} & \textbf{commonsense qa} & \textbf{arc easy} & \textbf{avg.}\\
        \midrule
        \midrule
        NoPE
        & \underline{25.67} & 43.71 & 51.86 & 29.80 & 76.70 & 41.83 & 68.83 & 31.61 & 51.40 & 42.14 \\
        ALiBi 
        & 24.97 & 42.53 & 53.12 & 31.40 & 77.70 & 43.06 & 69.42 & \textbf{33.42} & \textbf{53.68} & 42.93\\
        
        KERPLE 
        & 25.03 & \underline{43.81} & \textbf{54.07} & 32.40 & \textbf{78.20} & \underline{43.65} & 69.64 & 32.92 & 52.46 & \underline{43.22}\\

        FIRE
        & 25.60 & 42.63 & 49.88 & \underline{33.40} & 77.00 & 42.75 & 69.31 & 32.92 & 50.35 & 42.38 \\
        
        RoPE
        & 24.60 & 43.45 & 51.54 & \textbf{33.60} & 77.10 & 43.36 & \textbf{70.13} & \underline{33.01} & 52.98 & 42.98 \\

        FoPE
        & \textbf{26.17} & \textbf{44.12} & \underline{53.20} & 32.20 & \underline{77.80} & \textbf{43.83} & \underline{70.08} & 32.92 & \underline{53.33} & \textbf{43.37} \\
        
        \bottomrule
        \end{tabular}
    \end{adjustbox}
    \label{tab:downstream_acc}
\end{table*}

\begin{table*}[h]
    \centering
    \caption{Cross-Entropy Loss of 1.2B models with different positional embeddings on several downstream tasks.}
    \begin{adjustbox}{width=0.5\textwidth}
    \begin{tabular}{l|ccc|c}
        \toprule
        \textbf{Method} & \textbf{natural qs open} & \textbf{trivia qa wiki} & \textbf{arc easy} & \textbf{avg.}\\
        \midrule
        \midrule
        NoPE
        & 1.4334 & 1.6129 & 1.3077 & 1.4513 \\
        ALiBi 
        & \underline{1.3879} & 1.6057 & 1.2789 & 1.4242 \\
        
        KERPLE 
        & 1.4149 & \underline{1.5878} & 1.2825 & 1.4284 \\

        FIRE
        & 1.4365 & 1.6258 & 1.3118 & 1.4580 \\
        
        RoPE
        & 1.4114 & 1.5973 & \underline{1.2588} & \underline{1.4225} \\

        FoPE
        & \textbf{1.3818} & \textbf{1.5736} & \textbf{1.2272} & \textbf{1.3941} \\
        
        \bottomrule
        \end{tabular}
    \end{adjustbox}
    \label{tab:downsream_loss}
\end{table*}

\begin{table*}[h]
    \centering
    \caption{Evaluation of 1.2B models with different positional embeddings on mmlu.}
    \begin{adjustbox}{width=0.5\textwidth}
    \begin{tabular}{l|cccc|c}
        \toprule
        \textbf{Method} & \textbf{stem} & \textbf{social science} & \textbf{humanity} & \textbf{other} & \textbf{avg.}\\
        \midrule
        \midrule
        NoPE
        & 25.47 & 25.38 & \underline{28.01} & 24.39 & 25.81 \\
        ALiBi 
        & 25.93 & 27.20 & 26.44 & 24.39 & 25.99 \\
        
        KERPLE 
        & 26.02 & 25.97 & 27.82 & \underline{24.82} & 26.16 \\

        FIRE
        & 26.34 & \textbf{27.41} & 26.25 & 24.16 & 26.04 \\
        
        RoPE
        & \underline{27.01} & 27.26 & 27.87 & 24.53 & 26.68\\

        FoPE
        & \textbf{27.30} & \underline{27.31} & \textbf{29.89} & \textbf{25.79} & \textbf{27.57} \\
        
        \bottomrule
        \end{tabular}
    \end{adjustbox}
    \label{tab:mmlu}
\end{table*}

\section{Implementation Details}
\label{appendix:implementation}
\textbf{Implementation of FoPE} can be easily achieved with a weight matrix $W^F\in\mathbb{R}^{D \times (M-M_0)}$, where $M_0$ is the number of zero-frequency components in each head. This matrix maps the coefficients of all frequencies to a Fourier Series for each dimension. 
Since the zero-frequency sinusoidal function does not affect the original hidden states, the output dimension is less than the dimension of each head. 
To introduce more diversity and better simulate the randomness of the Spectrum Damage, we assign separate weights for different heads, as well as for the cosine and sine functions. In our implementation, gradients are not required for these matrices, so FoPE adds negligible memory and computation overhead compared to RoPE.

\textbf{Pseudo-code} of FoPE is shown in the final pages.
\lstset{ %
    language=Python,                 % 选择代码语言
    basicstyle=\ttfamily\footnotesize, % 设置基本字体样式
    numbers=left,                    % 显示行号
    numberstyle=\tiny\color{gray},   % 设置行号样式
    keywordstyle=\color{blue},       % 关键词颜色
    commentstyle=\color{gray},      % 注释颜色
    stringstyle=\color{gray},         % 字符串颜色
    breaklines=true,                 % 自动换行
    frame=single,                    % 使用方框
    tabsize=4,                       % 设置Tab大小
    captionpos=b                     % 设置标题位置
}

\begin{figure*}[ht]
\centering
\begin{lstlisting}
class FourierEmbedding(RotaryEmbedding):
    def __init__(self, config):
        super().__init__(config)

        self.input_dim = self.inv_freq.size(-1)
        self.output_dim = self.input_dim if self.input_dim <= self.head_dim//4 else self.head_dim//4

        self.sin_coef = nn.Parameter(
            torch.randn(self.config.n_heads, self.input_dim, self.output_dim),
            requires_grad=False
        )
        self.cos_coef = nn.Parameter(
            torch.randn(self.config.n_heads, self.input_dim, self.output_dim),
            requires_grad=False
        )
        torch.nn.init.xavier_normal_(self.sin_coef, gain=self.config.rope_fourier_init_norm_gain)
        torch.nn.init.xavier_normal_(self.cos_coef, gain=self.config.rope_fourier_init_norm_gain)

        self.sin_coef += torch.eye(self.input_dim, device=self.sin_coef.device)
        self.cos_coef += torch.eye(self.input_dim, device=self.cos_coef.device)
    
    def apply_rotary_pos_embed(self, pos_sin, pos_cos, t):
        fourier_sin = torch.einsum("bhtD, hDd -> bhtd", pos_sin, self.sin_coef / self.sin_coef.sum(dim=-2, keepdim=True))
        fourier_cos = torch.einsum("bhtD, hDd -> bhtd", pos_cos, self.cos_coef / self.cos_coef.sum(dim=-2, keepdim=True))

        fourier_sin = F.pad(input=fourier_sin, pad=(0, self.head_dim//2-fourier_sin.size(-1)), mode="constant", value=1)
        fourier_cos = F.pad(input=fourier_cos, pad=(0, self.head_dim//2-fourier_cos.size(-1)), mode="constant", value=1)

        fourier_sin = torch.cat((fourier_sin, fourier_sin), dim=-1)
        fourier_cos = torch.cat((fourier_cos, fourier_cos), dim=-1)

        return ((t * fourier_cos) - (self.rotate_half(t) * fourier_sin)).to(t.dtype)

class RotaryEmbedding(nn.Module):
    def __init__(self, config):
        super().__init__()
        self.config = config

        self.dim = self.config.d_model // self.config.n_heads
        self.inv_freq = self.get_inv_freq(self.dim)
   
   def get_inv_freq(self, dim):
        inv_freq = 1.0 / (
                self.config.rope_theta ** (torch.arange(0, dim, 2, device=device, dtype=torch.float) / dim)
        )
        if self.config.fope is True:
            inv_freq[inv_freq < 2*torch.pi/self.config.max_sequence_length] = 0
            inv_freq = inv_freq[inv_freq != 0.0]

        return inv_freq

\end{lstlisting}
\end{figure*}

\begin{figure*}[ht]
\centering
\begin{lstlisting}
    def get_rotary_embedding(self, seq_len):
        with torch.autocast(device.type, enabled=False):
            seq = torch.arange(seq_len, device=device, dtype=torch.float)
            freqs = torch.einsum("t, hd -> htd", seq, self.inv_freq)

            if self.config.fope is True:
                positions = freqs.unsqueeze(0)
            else:
                positions = torch.cat((freqs, freqs), dim=-1).unsqueeze(0)

        return positions.sin(), positions.cos()
        
    def rotate_half(self, x):
        B, nh, T, hs = x.size()
        x = x.view(B, nh, T, 2, hs // 2)
            
        x1, x2 = x.unbind(dim=-2)
        
        return torch.cat((-x2, x1), dim=-1)

    def apply_rotary_pos_emb(self, pos_sin, pos_cos, t):
        return ((t * pos_cos) - (self.rotate_half(t) * pos_sin)).to(t.dtype)

    def forward(self, x, all_len):
        with torch.autocast(x.device.type, enabled=False):
            x_len = x_.shape[-2]
            pos_sin, pos_cos = self.get_rotary_embedding(all_len)
            pos_sin = pos_sin.type_as(x_)
            pos_cos = pos_cos.type_as(x_)
            
            x_ = self.apply_rotary_pos_emb(
                pos_sin[:, :, all_len - x_len : all_len, :], 
                pos_cos[:, :, all_len - x_len : all_len, :], 
                x_,
            )
            
        return x_.type_as(x)
\end{lstlisting}
\end{figure*}

\section{Theoretical Details}
\label{appendix:theoretical}
\subsection{Derivation of Lemma \ref{lemma:harmonic}}
Given a non-linear function g(x), it can be rewritten as a power series by Taylor expansion:
\begin{equation}
    g(x)=\sum\limits_{p\in\mathbb{N}} a_px^p
\end{equation}
Suppose the input is a double-frequencies function:
\begin{equation}
    x(n)=\cos\omega_1n+\cos\omega_2n
\end{equation}
, the output becomes:
\begin{equation}\label{eq:fourier_power}
    g(x(n))=\sum\limits_{p\in\mathbb{N}} a_p(\cos\omega_1n+\cos\omega_2n)^p
\end{equation}
Considering the product conversion formula of sinusoid function:
\begin{equation}
    \cos\alpha\cos\theta=\frac{1}{2}[\cos(\alpha-\theta)+\cos(\alpha+\theta)]
\end{equation}
Thus, each sub-function of Eq.(\ref{eq:fourier_power}) can generate harmonic functions. For example, when $p=2$:
\begin{equation}
    \begin{aligned}
         ~(\cos\omega_1n+\cos\omega_2n)^2 =&~(\cos\omega_1n)^2+(\cos\omega_2n)^2+2\cos\omega_1n\cos\omega_2n \\
        =&~1+\frac{1}{2}\cos2\omega_1n+\frac{1}{2}\cos2\omega_2n+\cos(\omega_1-\omega_2)n+\cos(\omega_1+\omega_2)n
    \end{aligned}
\end{equation}

\subsection{Derivation of Equal (\ref{eq:undertrained})}
Before we begin the derivation, let's familiarize with two functions in time domain: \textbf{Rectangular Pulse Function} and \textbf{Single-Frequency Function}.

Given a \textbf{Rectangular Pulse} function: 
\begin{equation}
    r(t)=\left\{
    \begin{array}{cc}
         1, & |t|\leq T \\
         0, & |t|>T
    \end{array}
    \right.
\end{equation}
whose Fourier Transform is a \textbf{Sa Function}:
\begin{equation}
    R(\omega)=\int_{-T}^{T}e^{-j\omega t}dt=2\frac{\sin(\omega T)}{\omega}
\end{equation}

Given a \textbf{Single-Frequency} function:
\begin{equation}
    f(t)=e^{j\omega_m t}
\end{equation}
whose Fourier Transform is:
\begin{equation}
    F(\omega)=\int_{-\infty}^{+\infty}e^{j\omega_m t}e^{-j\omega t}dt
\end{equation}
which is equal to an \textbf{Impulse Function} in frequency domain:
\begin{equation}
    \delta(\omega_m)=\left\{
    \begin{array}{cc}
         1, & \omega = \omega_m \\
         0, & \omega \neq \omega_m
    \end{array}
    \right.
\end{equation}

Then, Eq (\ref{eq:undertrained}) can be easily derivated, as the context $x$ truncated with length $T$ can be seen as:
\begin{equation}
    x(t)=f(t)\cdot r(t)
\end{equation}
Its Fourier Transform in discrete case is Eq. (\ref{eq:undertrained}).

\subsection{Low-Pass Property of RoPE}
\label{appendix:lowpass}

As the frequency of RoPE obey $\omega_m=1/\theta^{(2m/M)}$, it only samples frequency less than 1 and samples more densely near 0.

According to the following Theorems and Corollary, RoPE only filters the low-frequency components from hidden states for attention, thus achieving a low-pass filter in attention.

\begin{theorem}
    For any continuous function, to reconstruct the original function using its discrete samples without distortion, the relation between the sampling frequency $f_s$ and the highest frequency component $f_h$ of the function must obey: $f_s \geq 2f_h$.
\end{theorem}

\begin{theorem}
    For any discrete function, the sampling frequency $\omega_s$ is constant: $\omega_s=2\pi$.
\end{theorem}

\begin{corollary}
    The frequency $\omega$ of any discrete function obey: $\omega \leq \pi$.
\end{corollary}

Comparing with high band $\omega\in(\pi/2, \pi]$, lower frequency components delivers a longer term influence on other tokens in time domain, letting attention have stronger bias to grab long-distance information.

\subsection{The Relationship Between the Theoretical Modeling and Practical Implementation of RoPE}

Let's define the query and key in the complex domain as $\mathbf{q}=||q||e^{i\theta_q}$ and $\mathbf{k}=||k||e^{i\theta_k}$, in the matrix domain as $\vv{\mathbf{q}}=[q_x, q_y]$ and $\vv{\mathbf{k}}=[k_x, k_y]$. According to the geometrical meaning of complex number, the relationships between the variations above are $||q||=\sqrt{q_x^2+q_y^2}$ and $\theta_q=\arctan\frac{q_x}{q_y} + k\pi$, with similar expressions holding for $||k||$ and $\theta_k$.

Thus, the implementation based on the rotary matrix reflects the geometric interpretation of phase rotation in the complex domain, as illustrated by the following derivation:

\begin{align}
&~~~~~\langle f_q(\mathbf{x_m},m),f_k(\mathbf{x_n},n)\rangle \\
&=\langle \mathbf{q}e^{im\theta},\mathbf{k}e^{in\theta}\rangle\\
&=Re[\mathbf{qk}^*e^{i(m-n)\theta}]\\
&=Re[\|q\|\|k\|e^{i[\theta_q-\theta_k+(m-n)\theta)]}]\\
&=\|q\|\|k\|\cos(\theta_q-\theta_k+(m-n)\theta)\\
&=\|q\|\|k\|\left[ \cos(\theta_q-\theta_k)\cos(m-n)\theta-\sin(\theta_q-\theta_k)\sin(m-n)\theta\right]\\
&=\|q\|\|k\|\left[(\cos\theta_q\cos\theta_k+\sin\theta_q\sin\theta_k)\cos(m-n)\theta 
-(\sin\theta_q\cos\theta_k-\cos\theta_q\sin\theta_k)\sin(m-n)\theta \right]\\
&=(q_xk_x+q_yk_y)\cos(m-n)\theta - (q_yk_x-q_xk_y)\sin(m-n)\theta\\
&=\begin{bmatrix}
q_x& q_y 
\end{bmatrix}
\begin{bmatrix}
\cos(m-n)\theta & -\sin(m-n)\theta  \\
\sin(m-n)\theta & \cos(m-n)\theta
\end{bmatrix}
\begin{bmatrix}
k_x  \\
k_y
\end{bmatrix}\\
&=\begin{bmatrix}
q_x& q_y 
\end{bmatrix}
\begin{bmatrix}
\cos m\theta & \sin m\theta  \\
-\sin m\theta & \cos m\theta
\end{bmatrix}
\begin{bmatrix}
\cos n\theta & -\sin n\theta  \\
\sin n\theta & \cos n\theta
\end{bmatrix}
\begin{bmatrix}
k_x  \\
k_y
\end{bmatrix}\\
&=(\mathbf{R}_{\theta,m}\vv{\mathbf{q}})^T(\mathbf{R}_{\theta,n}\vv{\mathbf{k}})\\
&=\vv{\mathbf{q}}^T\mathbf{R}_{\theta,m-n}\vv{\mathbf{k}}
\end{align}

%%%%%%%%%%%%%%%%%%%%%%%%%%%%%%%%%%%%%%%%%%%%%%%%%%%%%%%%%%%%%%%%%%%%%%%%%%%%%%%
%%%%%%%%%%%%%%%%%%%%%%%%%%%%%%%%%%%%%%%%%%%%%%%%%%%%%%%%%%%%%%%%%%%%%%%%%%%%%%%

\end{document}